\documentclass{article}
\usepackage[OT1]{fontenc}
\usepackage{iclr2027_conference,times}

\usepackage{amsmath,amsfonts,bm}

\def\eqref#1{equation~\ref{#1}}

\def\1{\bm{1}}

\def\vmu{{\bm{\mu}}}

\def\vc{{\bm{c}}}

\def\vp{{\bm{p}}}

\def\vx{{\bm{x}}}

\def\vz{{\bm{z}}}

\def\mI{{\bm{I}}}

\DeclareMathAlphabet{\mathsfit}{\encodingdefault}{\sfdefault}{m}{sl}
\SetMathAlphabet{\mathsfit}{bold}{\encodingdefault}{\sfdefault}{bx}{n}

\usepackage{amsmath}
\usepackage{amssymb}
\usepackage{algorithm}
\usepackage{algorithmic}
\usepackage{booktabs}
\usepackage{colortbl}
\usepackage{float}
\usepackage{graphicx}
\usepackage{hyperref}
\usepackage{pifont}
\newcommand{\cnum}[1]{\ding{\the\numexpr171+#1\relax}}
\usepackage[protrusion=false]{microtype}
\usepackage{url}

\title{WithinVAE: COMPACT VISUAL AUTOENCODING WITH SELF-DISTILLED SEMANTICS}

\author{%
Yuzhou Ji$^{1,2,*,\ddagger}$ \quad Jiepeng Wang$^{2,*}$ \quad Tianyu Jin$^{1}$ \quad Lizhuang Ma$^{1,\dagger}$\\
\bfseries Haibin Huang$^{2}$ \quad Chi Zhang$^{2}$ \quad Xuelong Li$^{2,\dagger}$\\[0.5em]
\normalfont $^{1}$Shanghai Jiao Tong University\\
\normalfont $^{2}$Institute of Artificial Intelligence, China Telecom\\[0.3em]
\normalfont\small $^{*}$Equal Contributions \quad $^{\dagger}$Corresponding Authors
}

\newcommand{\method}{WithinVAE}
\definecolor{tablegroupgray}{RGB}{235,235,235}
\definecolor{withinvaeblue}{RGB}{225,239,249}

\iclrfinalcopy
\hypersetup{
  pdftitle={WithinVAE: Compact Visual Autoencoding with Self-Distilled Semantics},
  pdfauthor={Yuzhou Ji, Jiepeng Wang, Tianyu Jin, Lizhuang Ma, Haibin Huang, Chi Zhang, Xuelong Li}
}

\begin{document}

\maketitle
\lhead{arXiv preprint}
\begingroup
\renewcommand{\thefootnote}{\fnsymbol{footnote}}
\footnotetext[3]{Work done during an internship at TeleAI}
\endgroup

\begin{abstract}
Visual autoencoders must balance reconstruction fidelity, compactness, and semantic structure. Existing semantic tokenizers typically inherit semantics from pretrained recognition encoders, whose invariances can discard the appearance detail required for inversion. We show that this inheritance is unnecessary: semantic structure can be learned inside an autoencoder's own reconstructive representation. We introduce \method, a compact variational visual autoencoder with a sequential bottleneck, DINO/iBOT-style self-distillation from an exponential-moving-average copy of the autoencoder itself, and position-dependent token masking to achieve a 4-fold compression ratio, which is free of pretrained representation encoder or external semantic target. On ImageNet-1K at $256\times256$, \method{} improves all four reported reconstruction metrics over RPiAE including PSNR by 1.36~dB, while using half its latent scalar capacity. Under frozen evaluation, the same latents improve linear probing by 6.5 points over RPiAE and 5.1 points over our reconstruction-only baseline, attributing the semantic gains to self-distillation rather than architecture or external supervision. An auxiliary latent-to-latent model reproduces the ordered-compression trend, and a pixel-to-latent model transfers the objective to a pretrained generative space. These results separate the origin of semantic supervision from its use: a compact visual autoencoder can organize semantic structure without borrowing a recognition representation.
\end{abstract}

\section{Introduction}
\label{sec:introduction}

\begin{figure}[b]
\centering
\includegraphics[width=\linewidth]{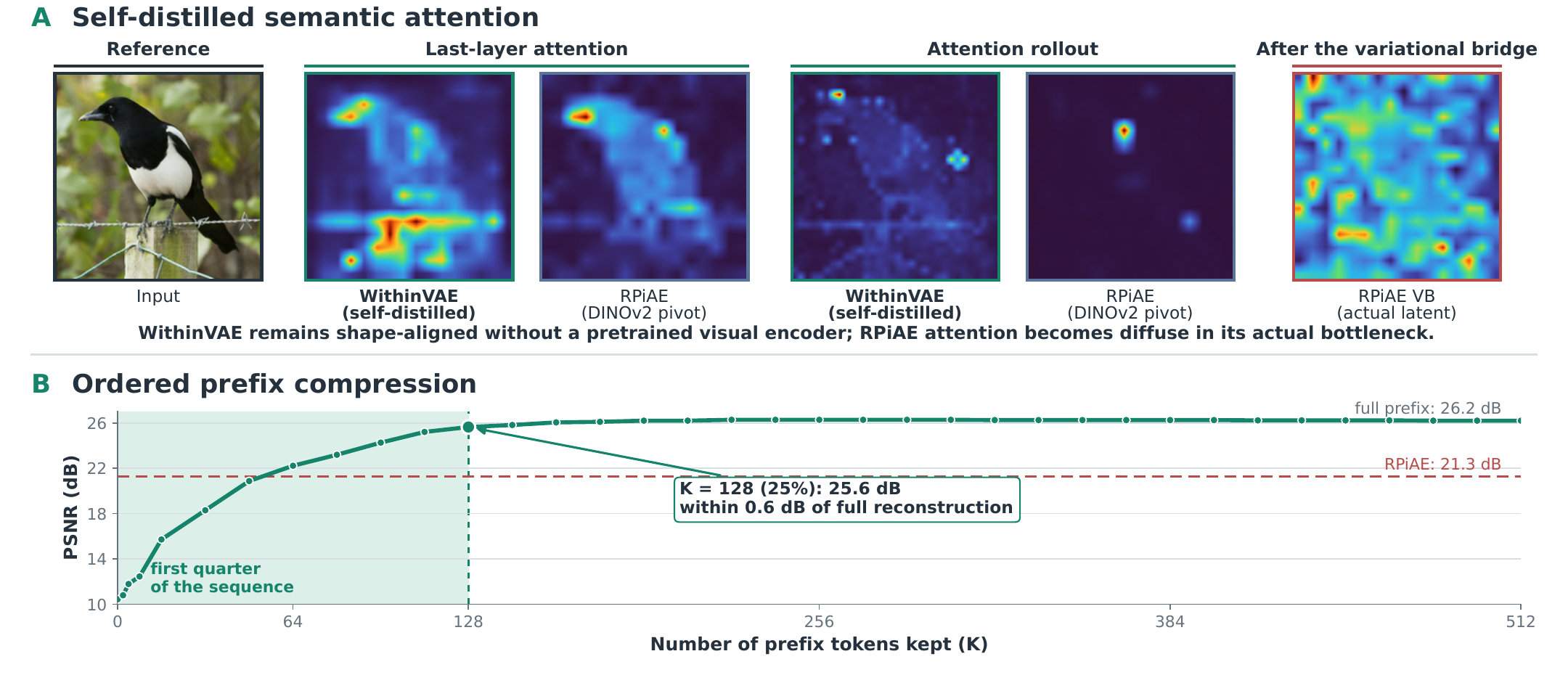}
\caption{Two diagnostics of self-distilled semantic autoencoding. \textbf{A:} Under matched display settings, \method{} attends to object extent without any pretrained recognition encoder, while RPiAE totally fails in providing reasonable attention after borrowed DINOv2 (rollout and actual latent after variational-bridge). \textbf{B:} Prefix decoding of the $512\times96$ latent-to-latent variant front-loads reliability: $K=128$ retains 25.6~dB PSNR, within 0.6~dB of the full sequence, and $K=64$ (6,144 scalars, under half of RPiAE's budget) already exceeds RPiAE's 21.3~dB.}
\label{fig:teaser}
\end{figure}

Latent generative models replace pixels with a lower-dimensional model-facing representation. The tokenizer that defines this interface determines reconstruction fidelity, latent dimensionality, and the geometry seen by every downstream generator. Ideally, its latents preserve visual detail, expose semantic structure, and remain compact. These requirements are not automatically compatible: stronger compression removes information, pixel losses favor local appearance, and recognition-oriented invariance can suppress precisely the variation needed for reversible generation~\citep{shannon1959coding,blau2019rethinking,xi2026survey,li2024positive,zhang2025variational}.

Recent semantic tokenizers address this tension by transferring representations from pretrained vision encoders. Some align autoencoding or diffusion features to DINO- or CLIP-like targets~\citep{yu2024repa,xu2025reals,yao2025vavae,zhang2025psvae}; others decode, compress, or adapt the representation encoder itself~\citep{zheng2025rae,singh2026raev2,gong2026rpiae,gao2025fae,chen2026aligntok,shi2026svg}. These methods demonstrate the value of semantics, but they also couple the tokenizer to an understanding model whose invariances were learned for a different objective. Freezing the encoder restricts reconstruction adaptation, whereas unconstrained fine-tuning can erase the inherited geometry. RPiAE addresses this compromise with a frozen pivot and a variational bridge~\citep{gong2026rpiae}. We study the complementary question: \emph{must the semantic target come from an external representation model at all?}

Our hypothesis is that view consistency and masked prediction can organize semantics within the reconstructive representation itself. One-dimensional tokenizers provide a sequence axis on which nonuniform corruption can order information~\citep{yu2024titok,miwa2025onedpiece}, while EMA teacher--student objectives organize global and patch representations without labels~\citep{caron2021dino,zhou2022ibot}. Their combination need not make compression lossless or guarantee semantic ordering. It instead creates two coupled pressures: early tokens are more reliable for reconstruction, and the latent space must remain consistent across appearance-changing views.

We instantiate this idea in \method, a compact image-to-image variational autoencoder initialized from the generative One-D-Piece architecture~\citep{miwa2025onedpiece}. It maps an image to a one-dimensional latent token sequence, applies a token-wise variational bottleneck, and reconstructs pixels. During training, token masking probability increases with sequence position. A separate EMA copy supplies global DINO-style targets from pooled compact tokens and local iBOT-style targets on masked image patches. ``Self-distilled semantics'' refers specifically to the origin of these targets: no DINO, DINOv2, or CLIP weights supervise training. It does \emph{not} mean training every generative component from scratch.

The experiments establish three results. First, at half of RPiAE's latent scalar capacity, \method{} improves all four reported reconstruction metrics simultaneously (rFID, PSNR, LPIPS, SSIM), and its 128-token prefix remains stronger than RPiAE in PSNR and SSIM at one quarter of that capacity. Second, frozen linear-probe and nearest-neighbor evaluations attribute the semantic gains to the self-distillation objective itself: it improves linear probing by 5.072 points over our reconstruction-only baseline and by 6.482 points over RPiAE, with simultaneous nearest-neighbor gains ruling out collapse. Third, the representation is directly modelable by a latent diffusion transformer---a 675M LightningDiT trained on \method{} latents already exceeds RPiAE on recall without guidance---although the remaining generation metrics trail RPiAE, which we analyze as a generator--autoencoder compatibility question rather than a failure of the representation. Our claims are accordingly improved reconstruction at lower capacity, capacity-efficient prefix decoding, and frozen semantic quality independent of external recognition features; state-of-the-art generation is not among them.

Our contributions are:
\begin{itemize}
    \item We formulate \emph{self-distilled semantic autoencoding}, in which semantic supervision is produced online by the autoencoder's own EMA copy rather than transferred from a pretrained recognition encoder.
    \item We instantiate it as a compact variational autoencoder coupling position-dependent token masking with global--local EMA self-distillation, and show it improves all four reconstruction metrics over RPiAE at half the scalar capacity (+1.36~dB PSNR on $256\times32$ model), while its 128-token prefix still exceeds RPiAE in PSNR and SSIM at one quarter.
    \item We attribute the semantic gains through controlled component ablations (+5.1 linear-probe points over the reconstruction-only baseline, with reconstruction improving at the same time), and verify that the objective transfers across pixel and pretrained generative-latent interfaces, including prefix decoding that retains 25.6~dB PSNR at $4\times$ fewer tokens without retraining.
\end{itemize}

\section{Related Work}
\label{sec:related}

Visual generation and understanding span image, video, and structured 3D representations~\citep{xi2026survey,an2026ai,wang2025mmgen,xi2025ctrlvdiff,xi2026omnivdiff,wang2023structre,dong2025crossgen}, with compression placing additional demands on compact representations~\citep{chen2026generative}.

\paragraph{Visual tokenization and compact sequences.}
VQ-VAE and VQGAN established discrete image codes for generative modeling~\citep{oord2017neural,esser2021vqgan}, while latent diffusion popularized continuous spatial latents~\citep{rombach2022ldm}. Modern tokenizers improve codebooks, architectures, and image--video unification~\citep{yu2022vector,luo2024openmagvit2,weber2024maskbit,wang2024omnitokenizer}. A complementary line removes the fixed two-dimensional grid. SEED learns compact tokens for multimodal models~\citep{ge2024seed}, TiTok compresses images with latent queries~\citep{yu2024titok}, and One-D-Piece orders a one-dimensional sequence with Tail Token Drop~\citep{miwa2025onedpiece}. Rateless autoencoding similarly removes later dimensions stochastically~\citep{koikeakino2020stochastic}. We retain the query-based sequence topology, but use continuous variational tokens and a position-dependent reliability profile. Our method achieves excellent compression as reported in token count, width, and total scalar count, while maintaining reconstruction quality.

\paragraph{Representation priors for generation.}
Self-supervised and vision--language encoders provide useful semantic geometries~\citep{radford2021learning,oquab2024dinov2}. REPA and REPA-E align diffusion features without replacing the tokenizer~\citep{yu2024repa,leng2025repae}; ReaLS, VA-VAE, and PS-VAE align the autoencoding latent~\citep{xu2025reals,yao2025vavae,zhang2025psvae}. RAE decodes frozen representation features, FAE compresses them, SVG separates semantic and residual features, and AlignTok progressively adapts a foundation encoder~\citep{zheng2025rae,singh2026raev2,gao2025fae,shi2026svg,chen2026aligntok,shao2026aiflow}. RPiAE fine-tunes a representation-initialized encoder under a frozen pivot before learning a compact variational bridge~\citep{gong2026rpiae}. \method{} differs in where semantic supervision originates: its targets are produced online by the autoencoder's EMA copy.

\paragraph{Self-supervision inside an autoencoder.}
Controlled corruption can also benefit representation learning~\citep{li2024positive,zhang2025variational}. Masked autoencoding learns by reconstructing hidden content~\citep{he2022masked}; DINO uses centered and sharpened EMA targets across views~\citep{caron2021dino}; iBOT extends this objective to masked patches~\citep{zhou2022ibot}. MAETok combines masked reconstruction with external feature targets for diffusion-oriented tokens~\citep{chen2025maetok}. We borrow the optimization forms of DINO and iBOT, not their pretrained representations. The global branch directly supervises pooled compact tokens, whereas the local branch regularizes the shared image-patch encoder. This distinction matters when local masked prediction can shape the bottleneck through the shared encoder, but is not itself a loss on individual compact tokens.

\section{Method}
\label{sec:method}

\begin{figure}[t]
\centering
\includegraphics[width=\linewidth]{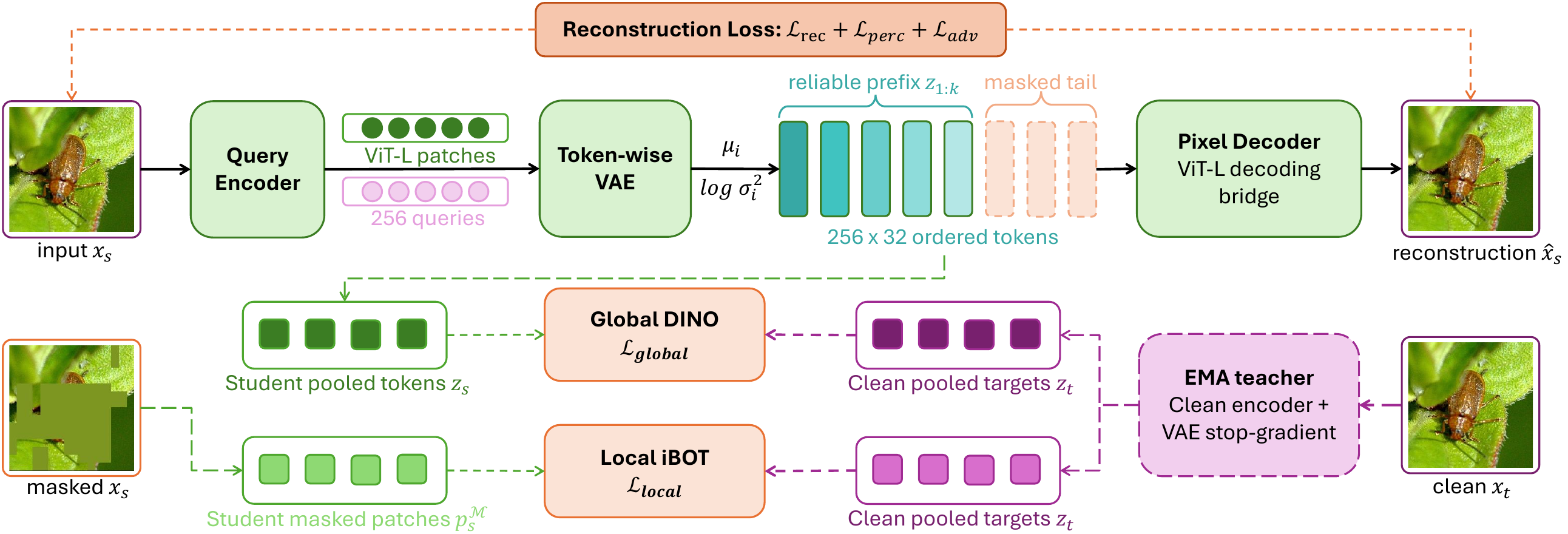}
\caption{\method{} training. The reconstruction path maps an image $x_s$ through a query encoder (ViT-L patches and 256 latent queries) and a token-wise VAE to a $256\times32$ ordered token sequence; position-dependent masking keeps a reliable prefix $z_{1:k}$ and corrupts the masked tail before the ViT-L pixel decoder produces $\hat{x}_s$, supervised by a reconstruction loss $\mathcal{L}_{\rm rec}+\mathcal{L}_{\rm perc}+\mathcal{L}_{\rm adv}$. The semantic branches below match mean-pooled compact tokens via global DINO ($\mathcal{L}_{\rm global}$) and student masked patches via local iBOT ($\mathcal{L}_{\rm local}$) against targets produced by an EMA copy of the encoder and VAE that observes a clean view $x_t$.}
\label{fig:method}
\end{figure}

\subsection{WithinVAE: compact variational autoencoder}

Figure~\ref{fig:method} gives the complete training system. A single encoder and bottleneck feed two paths that meet only through shared weights: an upper reconstruction path that must invert the image from an ordered, partially corrupted token sequence, and a lower semantic path that matches student outputs against an EMA copy of the same autoencoder. The subsections below specify these paths in turn.

Given an image $\vx\in[0,1]^{3\times H\times W}$, an encoder $E_\theta$ aggregates image patches with $N$ learned queries, a bottleneck $B_\psi$ produces continuous tokens $\vz\in\mathbb{R}^{N\times d}$, and a decoder $D_\phi$ reconstructs $\hat{\vx}$. Our main model uses $N=256$, $d=32$, and ViT-L encoder and decoder backbones. This is the reconstruction path along the top of Figure~\ref{fig:method}. A diagonal posterior is predicted independently at each token:
\begin{equation}
(\vmu_i,\log\bm{\sigma}_i^2)=B^{\mathrm{enc}}_\psi(E_\theta(\vx)_i),\qquad
\vz_i=\vmu_i+\bm{\sigma}_i\odot\bm{\epsilon}_i,
\quad \bm{\epsilon}_i\sim\mathcal{N}(\mathbf{0},\mI).
\label{eq:posterior}
\end{equation}
The posterior mean is used at evaluation. Training combines pixel L1, ConvNeXt perceptual, KL, and generator-adversarial terms with a LeCam-regularized discriminator. Adding semantic self-distillation losses gives
\begin{equation}
\mathcal{L}=\mathcal{L}_{\mathrm{rec}}+\lambda_{\mathrm{perc}}\mathcal{L}_{\mathrm{perc}}
+\lambda_{\mathrm{adv}}\mathcal{L}_{\mathrm{adv}}+\lambda_{\mathrm{KL}}\mathcal{L}_{\mathrm{KL}}
+\lambda_{\mathrm{g}}\mathcal{L}_{\mathrm{global}}+\lambda_{\mathrm{l}}\mathcal{L}_{\mathrm{local}}.
\label{eq:full_objective}
\end{equation}
The KL term is averaged over tokens and summed over latent dimensions. Shape-compatible encoder, decoder, latent-query, and pixel-decoder weights are initialized from One-D-Piece; the 32-dimensional boundary projections and the variational bottleneck are random. Thus our comparison isolates external \emph{semantic} supervision, not generative pretraining.

\subsection{Position-dependent token reliability}
\label{sec:progressive_mask}

At training time, reconstruction uses independently masked tokens
\begin{equation}
p_i=p_{\max}\frac{i-1}{N-1},\qquad
m_i\sim\mathrm{Bernoulli}(p_i),\qquad
\tilde{\vz}_i=(1-m_i)\vz_i,
\label{eq:progressive_mask}
\end{equation}
with $p_{\max}=0.5$. The increasing corruption probability makes early tokens statistically more reliable than late tokens, biasing the encoder to place broadly useful information near the front of the sequence. This is the split drawn between the reliable prefix $z_{1:k}$ and the corrupted tail in Figure~\ref{fig:method}: only the surviving sequence reaches the pixel decoder. We test this predicted reliability ordering directly by decoding variable-length prefixes,
\begin{equation}
\hat{\vx}_K=D_\phi(\vz_{1:K}),\qquad 1\leq K\leq N,
\label{eq:prefix}
\end{equation}
using the decoder's variable-length cross-attention. No retraining is performed for a new $K$.

\subsection{Global and local semantic self-distillation}

Let $f_\theta$ denote the student encoder, bottleneck, mean pool, and prototype head, and $f_{\bar\theta}$ its stop-gradient EMA copy, shown as the teacher branch at the lower right of Figure~\ref{fig:method}. Two views $\vx_s$ and $\vx_t$ of the same image produce
\begin{equation}
\vp_s=\mathrm{softmax}(f_\theta(\vx_s)/\tau_s),\qquad
\vp_t=\mathrm{softmax}((f_{\bar\theta}(\vx_t)-\vc)/\tau_t),
\quad
\mathcal{L}_{\mathrm{global}}=-\sum_j\vp_{t,j}\log\vp_{s,j}.
\label{eq:global_dino}
\end{equation}
The head predicts 8,192 normalized prototypes from the mean of the same compact tokens passed to the reconstruction decoder before token masking. Centering, teacher sharpening, and an EMA momentum scheduled from 0.996 toward 1 discourage collapse.

For local structure, block masks cover 10--50\% of the encoder's $16\times16$ image-patch grid. The student observes a mean-filled masked view and the EMA teacher the corresponding clean view. For masked patch set $\mathcal{M}$,
\begin{equation}
\mathcal{L}_{\mathrm{local}}=-\frac{1}{|\mathcal{M}|}
\sum_{u\in\mathcal{M}}\sum_j\vp^{(u)}_{t,j}\log\vp^{(u)}_{s,j}.
\label{eq:local_ibot}
\end{equation}
The local branch reuses the encoder and prototype head but bypasses the compact-token mean pool, which is why Figure~\ref{fig:method} draws $\mathcal{L}_{\rm local}$ from the student masked patches rather than from the token sequence. Its role is to preserve spatially grounded patch information in the shared encoder while the global branch organizes the compact bottleneck.

\subsection{Complementary roles of the objectives}

The progressive mask and the two semantic losses in Equation~\ref{eq:full_objective} act at different interfaces. Position-dependent corruption changes which compact tokens are reliable to the pixel decoder, but supplies no semantic label and does not by itself require view invariance. Global self-distillation instead compares mean-pooled compact sequences across two augmented views. It directly regularizes the representation exposed to a downstream model, but pooling can discard spatial evidence. Local masked prediction supplies that missing pressure at image-patch resolution through the shared encoder. It therefore affects compact tokens only indirectly and should not be interpreted as assigning a semantic label to every latent position.

The EMA teacher is also materially different from a fixed representation pivot. Its targets co-evolve with the reconstructive model, while stop-gradient, centering, teacher sharpening, and delayed loss warm-up stabilize the teacher--student loop. Reconstruction prevents a purely invariant solution from ignoring image detail; self-distillation discourages the reconstruction space from becoming useful only to its paired decoder. This coupling motivates the central empirical questions: whether semantic gains survive frozen evaluation, whether prefix reliability appears despite independent rather than contiguous masking, and whether the resulting interface remains usable by a separately trained generator.

\section{Experiments}
\label{sec:experiments}

\subsection{Implementation details}

We organize the evaluation around four questions. First, does semantic self-distillation preserve reconstruction while reducing the model-facing capacity? Second, are any semantic gains visible to frozen, decoder-independent probes rather than only to the training heads? Third, does the position-dependent reliability profile support prefix decoding? Finally, can the same representation be modeled by a diffusion transformer and can the objective transfer across reconstruction interfaces? Each claim below is tied to the corresponding metric; no single score is used as evidence for all four questions.

\paragraph{Data and optimization.}
We train on ImageNet-1K at $256\times256$ with random crops and horizontal flips. Reconstruction uses L1, ConvNeXt-S perceptual, KL, and adversarial weights of 1, 0.1, $3.75\times10^{-5}$, and 0.01; global and local distillation each use 0.05. KL, adversarial, and semantic terms use scheduled warm-up, and the teacher temperature increases from 0.04 to 0.07. AdamW uses a per-GPU batch size of 16 with four-step gradient accumulation, learning rate and weight decay $10^{-4}$, $(\beta_1,\beta_2)=(0.9,0.999)$, gradient clipping at 1.0, and bfloat16.

\paragraph{Architecture and initialization.}
The $16\times16$ image-patch grid and 256 learned latent queries are processed by a ViT-L encoder. A token-wise MLP of hidden width 2,048 predicts the 32-dimensional posterior parameters, and a ViT-L decoder maps the sampled sequence back to pixel patches. This yields a nonspatial model interface: the decoder addresses a sequence of compact tokens rather than a fixed $h\times w$ latent grid. At evaluation we use posterior means, so stochastic sampling affects training regularization but not the reported reconstructions or frozen features.

Shape-compatible encoder, decoder, query, and pixel-head parameters are initialized from One-D-Piece. Boundary projections introduced by the narrower latent width, the variational bottleneck, and the 8,192-prototype semantic head are new. Pretrained parameters use a $0.1\times$ learning-rate multiplier, while new parameters use the full rate. The EMA teacher duplicates the autoencoder-side encoder and semantic pathway; it neither loads nor queries an external recognition network. Consequently, the experiment removes recognition-feature supervision while retaining a generative architectural initialization, a distinction maintained throughout our claims.

\paragraph{Reconstruction and capacity.}
All \method{} reconstruction metrics use all 50,000 ImageNet validation images under deterministic preprocessing and lossless image export. We report rFID, PSNR, LPIPS-Alex, and SSIM. For continuous latents, ``scalars'' means token count times channel width. It is a transparent proxy for the interface presented to a downstream model, not an entropy-coded bitrate. Baseline values are taken from the cited papers; our local RPiAE reconstruction gives 20.75 dB PSNR, consistent with rather than stronger than its reported 21.3 dB.

\paragraph{Frozen semantic evaluation.}
We mean-pool compact tokens and freeze each autoencoder. Linear probing trains one classifier with SGD for 100 epochs. For 20-NN, L2-normalized features use exact cosine neighbors and temperature-weighted voting. Same-image view consistency is $100$ times cosine similarity between two independently cropped and flipped validation views, reported as mean $\pm$ standard deviation. Consistency is interpreted only alongside the non-collapsed linear and nearest-neighbor results.

\paragraph{Training duration and compute.}
The main autoencoder is trained for 10 epochs, and the downstream LightningDiT is trained for 80 epochs. All reported experiments are conducted on eight NVIDIA H100 GPUs. Reconstruction, frozen-semantic, prefix, and attention evaluations use the trained autoencoder and its EMA teacher; generation samples classes uniformly and decodes generated latents with the corresponding decoder.

\subsection{Quantitative results}

Table~\ref{tab:reconstruction} reports the main comparison. Against RPiAE---the closest baseline pursuing the same compact, one-dimensional, semantic, variational interface---\method{} halves scalar capacity from 16,384 to 8,192 while improving all four reconstruction metrics simultaneously: rFID 0.46 vs.\ 0.50, PSNR +1.36~dB, LPIPS $-0.057$, and SSIM $+0.068$. The 128-token prefix uses only 4,096 scalars, one quarter of RPiAE's budget, and still exceeds RPiAE's full-sequence result in PSNR and SSIM. For context, SD-VAE and VA-VAE use spatial ($h\!\times\!w$ grid) latents and reach higher PSNR at comparable budgets, but they are two-dimensional interfaces and VA-VAE depends on a pretrained representation model; the comparison that controls for interface type and supervision origin is the one against RPiAE.

\begin{table}[t]
\caption{ImageNet-1K reconstruction at $256\times256$. \method{} improves all four metrics over RPiAE at half its scalar count, without any external representation model. Baselines are reported values from their source papers (consolidated in RPiAE Table 3); \method{} uses our 50K-image evaluation. Ext. RM indicates an external pretrained representation model used for semantic initialization or supervision. Scalar count is an interface-capacity proxy, not a bitrate.}
\label{tab:reconstruction}
\centering
\small
\setlength{\tabcolsep}{4.2pt}
\begin{tabular}{lccrrrrr}
\toprule
Model & Ext. RM & Latent & Scalars & rFID$\downarrow$ & PSNR$\uparrow$ & LPIPS$\downarrow$ & SSIM$\uparrow$ \\
\midrule
SD-VAE~\citep{rombach2022ldm} & No & $32\!\times\!32\!\times\!4$ & 4,096 & 0.61 & 26.9 & 0.130 & 0.736 \\
VA-VAE~\citep{yao2025vavae} & Yes & $16\!\times\!16\!\times\!32$ & 8,192 & 0.27 & 27.7 & 0.097 & 0.779 \\
RAE-S~\citep{zheng2025rae} & Yes & $256\!\times\!384$ & 98,304 & 0.64 & 18.9 & 0.252 & 0.489 \\
RAE-B~\citep{zheng2025rae} & Yes & $256\!\times\!768$ & 196,608 & 0.57 & 18.8 & 0.256 & 0.483 \\
FAE-d32~\citep{gao2025fae} & Yes & $16\!\times\!16\!\times\!32$ & 8,192 & 0.68 & -- & -- & -- \\
RPiAE~\citep{gong2026rpiae} & Yes & $256\!\times\!64$ & 16,384 & 0.50 & 21.3 & 0.216 & 0.525 \\
\rowcolor{withinvaeblue}
\textbf{\method{}} & No & $256\!\times\!32$ & 8,192 & \textbf{0.46} & \textbf{22.66} & \textbf{0.159} & \textbf{0.593} \\
\rowcolor{withinvaeblue}
\textbf{\method{} prefix} & No & $128\!\times\!32$ & \textbf{4,096} & 0.47 & 22.33 & 0.163 & 0.581 \\
\bottomrule
\end{tabular}
\end{table}

The trend is not limited to one prefix. Figure~\ref{fig:teaser}B reports the denser sweep for the auxiliary latent-to-latent model: its $64\times96$ prefix uses 6,144 scalars and reaches 22.23 dB PSNR, while $128\times96$ reaches 25.63 dB, within 0.64 dB of the full sequence. The progressive mask therefore yields useful prefix inference despite never training with contiguous tail removal.

\subsection{Ablation study}

Table~\ref{tab:semantics} separates architectural and semantic effects; circled indices refer to its rows. The compact architecture accounts for part of the frozen-probe gap on its own: \cnum{2} exceeds RPiAE \cnum{1} by 1.410 linear-probe points (66.814\% vs.\ 65.404\%) with no semantic objective. Each component then adds further---progressive masking \cnum{3} adds 0.998 over \cnum{2}, the global objective \cnum{4} adds 1.982 over \cnum{3}, and the local objective \cnum{5} adds 2.500---and the two semantic branches are not redundant, since combining them in \cnum{6} gains 1.574 over \cnum{5} and 2.092 over \cnum{4} to reach 71.886\%, i.e.\ 5.072 above \cnum{2} and 6.482 above \cnum{1}.

The intermediate rows show that fidelity and semantics improve together only for the complete objective. Taken alone, masking and the global branch cost reconstruction: \cnum{3} and \cnum{4} fall 0.21 dB and 0.43 dB below \cnum{2} in PSNR. The local branch \cnum{5} recovers fidelity, and only \cnum{6} is best on both axes at once (22.66 dB, 0.159 LPIPS; $+0.76$~dB and $-0.016$ over \cnum{2}) while adding the 5.072-point probe gain. Semantic organization is therefore not purchased with reconstruction quality, even though two of the three components would individually suggest such a trade-off. Collapse is also ruled out: \cnum{6} improves 20-NN by 5.538 points over \cnum{1} and 2.959 over \cnum{3}, so its 99.125 view score is accompanied by---not substituted for---class-discriminative structure.

\begin{table}[t]
\caption{Reconstruction and frozen-latent semantics on ImageNet-1K. The full objective improves both fidelity and linear-probe accuracy over every partial variant. Circled indices \cnum{1}--\cnum{6} label rows for reference in the text. RPiAE reconstruction values are published; its semantic columns use our frozen-latent protocol. LP and 20-NN are top-1 accuracy (\%). View is same-image cross-view cosine multiplied by 100. Dashes denote evaluations not run for intermediate ablations.}
\label{tab:semantics}
\centering
\small
\setlength{\tabcolsep}{5.0pt}
\begin{tabular}{clrrrrr}
\toprule
& Variant & PSNR$\uparrow$ & LPIPS$\downarrow$ & LP$\uparrow$ & 20-NN$\uparrow$ & View$\uparrow$ \\
\midrule
\cnum{1} & RPiAE & 21.30 & 0.216 & 65.404 & 66.856 & $98.339\!\pm\!1.466$ \\
\cnum{2} & \method{} reconstruction only & 21.90 & 0.175 & 66.814 & -- & -- \\
\cnum{3} & $+$ progressive mask & 21.69 & 0.180 & 67.812 & 69.435 & $98.029\!\pm\!1.441$ \\
\cnum{4} & $+$ mask $+$ global & 21.47 & 0.186 & 69.794 & -- & -- \\
\cnum{5} & $+$ mask $+$ local & 22.13 & 0.162 & 70.312 & -- & -- \\
\rowcolor{withinvaeblue}
\cnum{6} & \textbf{\method{} full} & \textbf{22.66} & \textbf{0.159} & \textbf{71.886} & \textbf{72.394} & $\mathbf{99.125\!\pm\!0.732}$ \\
\bottomrule
\end{tabular}
\end{table}

Figure~\ref{fig:teaser}A provides a controlled attention diagnostic. The displays hold the input crop, preprocessing, per-map normalization, and color map fixed, but the attention operators are necessarily architecture-specific. The \method{} map averages latent-query-to-patch attention across heads at encoder layer 10 (zero-based) of the EMA teacher. The RPiAE map averages its variational-bridge encoder attention over heads and bridge queries. Figure~\ref{fig:qualitative}A extends the comparison to seven matched validation images. We do not use attention as proof of causal token contribution; the quantitative semantic claim rests on the frozen evaluations above.

\subsection{Downstream generation}

We train a 675M-parameter LightningDiT~\citep{yao2025vavae} for 80 epochs directly on \method{} latents; Table~\ref{tab:generation} confirms that the representation is immediately modelable by a diffusion transformer. Without guidance, \method{} already exceeds RPiAE on recall (0.66 vs.\ 0.60), indicating that the compact sequence preserves sample diversity. The remaining generation metrics trail RPiAE (e.g., gFID 9.12 vs.\ 2.25 unguided), and guidance narrows but does not close the gap: better autoencoder reconstruction and frozen semantics do not automatically yield a better generator. We view this as a latent-normalization and generator--autoencoder compatibility question rather than a failure of the representation, and as the clearest direction for future work.

\begin{table}[t]
\caption{Class-conditional ImageNet generation with a 675M LightningDiT trained for 80 epochs. RPiAE values are reported by \citet{gong2026rpiae}; \method{} uses a 50K-sample evaluation. Guidance settings follow each implementation and are not asserted to be identical beyond the reported on/off condition.}
\label{tab:generation}
\centering
\small
\begin{tabular}{llrrrr}
\toprule
Model & Guidance & gFID$\downarrow$ & IS$\uparrow$ & Prec.$\uparrow$ & Rec.$\uparrow$ \\
\midrule
RPiAE & Off & 2.25 & 208.7 & 0.81 & 0.60 \\
RPiAE & On & 1.51 & 225.9 & 0.79 & 0.65 \\
\rowcolor{withinvaeblue}
\method{} & Off & 9.12 & 124.3 & 0.69 & \textbf{0.66} \\
\rowcolor{withinvaeblue}
\method{} & On & 3.67 & 169.2 & 0.76 & 0.58 \\
\bottomrule
\end{tabular}
\end{table}

\subsection{Qualitative results}

\begin{figure}[t]
\centering
\includegraphics[width=0.90\linewidth]{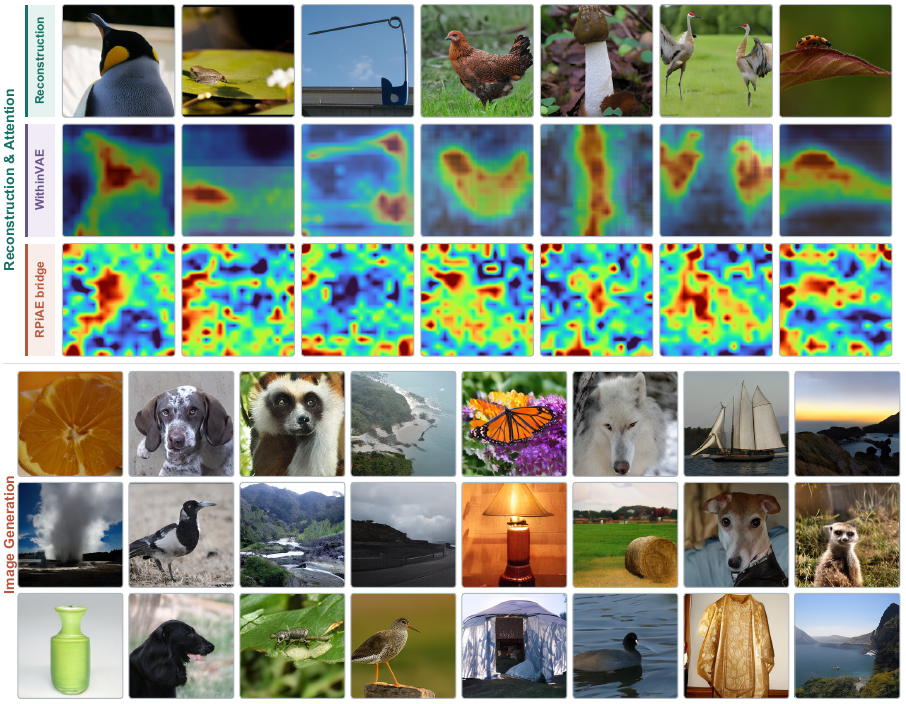}
\caption{Qualitative results. \textbf{Up:} Matched validation examples showing \method{} reconstruction, \method{} head-averaged attention, and RPiAE variational-bridge attention averaged over heads and bridge queries. The displays share the input crop, per-map normalization, and color map, but the attention operators are architecture-specific. \textbf{Down:} Class-conditional generation samples from the 80-epoch LightningDiT. Table~\ref{tab:generation} provides the distribution-level comparison.}
\label{fig:qualitative}
\end{figure}

Figure~\ref{fig:qualitative} compares the main model against RPiAE on seven matched validation images. In the upper panel---including a thin safety pin, multiple cranes, and a small insect---\method{} attention concentrates on the foreground more clearly than the RPiAE bridge maps, which are more fragmented. Both remain architecture-specific, coarse diagnostics that can extend into correlated background regions; neither is interpreted as segmentation or causal evidence. The lower panel shows recognizable object and scene structure across animal, object, and landscape classes, but a selected montage cannot expose coverage failures or the frequency of malformed samples, so the 50K-sample metrics in Table~\ref{tab:generation} remain the basis for comparing generators.

\subsection{Transfer across reconstruction interfaces}

We test the training principle in two auxiliary systems: pixels compressed to a pretrained Qwen-Image latent, and Qwen-Image latents compressed and reconstructed in their own space~\citep{wu2025qwenimage}. Table~\ref{tab:generality} reports end-to-end RGB metrics after the Qwen-Image decoder. These models do not share the main architecture and their full $512\times96$ bottlenecks are larger, so they test transfer of the objective rather than superiority of one architecture or budget. The transfer succeeds on the key phenotype: the latent-to-latent model preserves the prefix trend, losing only 0.64 dB from 512 to 128 tokens---evidence that ordered reliability follows from the objective rather than from one architecture.

The two interfaces answer different questions. Pixel-to-Qwen-latent reconstruction must learn both compression and the geometry expected by a fixed generative decoder, so its 23.91 dB PSNR cannot be attributed solely to the self-distillation objective. Qwen-latent-to-Qwen-latent reconstruction removes that first mapping and reaches 26.27 dB, but exposes 49,152 scalars at full length; its prefix curve is the informative result, since a $4\times$ sequence reduction moves rFID only from 0.40 to 0.42 and PSNR from 26.27 to 25.63 dB without retraining. At 64 tokens the sharper PSNR and SSIM drop shows that ordered reliability is useful but not lossless.

\begin{table}[t]
\caption{End-to-end reconstruction across interfaces at $256\times256$. Qwen rows use a pretrained generative autoencoder; metrics are computed in RGB after decoding. Scalars expose the unequal model-facing capacities.}
\label{tab:generality}
\centering
\small
\setlength{\tabcolsep}{4.0pt}
\begin{tabular}{lrrrrrr}
\toprule
Interface & Bottleneck & Scalars & rFID$\downarrow$ & PSNR$\uparrow$ & LPIPS$\downarrow$ & SSIM$\uparrow$ \\
\midrule
Pixel $\rightarrow$ pixel & $256\!\times\!32$ & 8,192 & 0.46 & 22.66 & 0.159 & 0.593 \\
Pixel $\rightarrow$ pixel, prefix & $128\!\times\!32$ & 4,096 & 0.47 & 22.33 & 0.163 & 0.581 \\
Pixel $\rightarrow$ Qwen latent & $512\!\times\!96$ & 49,152 & 0.43 & 23.91 & 0.144 & 0.614 \\
Qwen latent $\rightarrow$ Qwen latent & $512\!\times\!96$ & 49,152 & 0.40 & 26.27 & 0.131 & 0.714 \\
\quad prefix & $128\!\times\!96$ & 12,288 & 0.42 & 25.63 & 0.136 & 0.698 \\
\quad prefix & $64\!\times\!96$ & 6,144 & 0.48 & 22.23 & 0.160 & 0.579 \\
\bottomrule
\end{tabular}
\end{table}

\subsection{Analysis of Experimental Results}

The ablations, the transfer results, and the generation result answer three different questions, and only together do they identify what is responsible for the gains. The generator result is the necessary negative control: frozen classification and reconstruction quality do not guarantee an easier diffusion distribution. The central claim therefore concerns the source of semantic supervision---a pretrained recognition encoder is unnecessary for the measured semantic organization. Table~\ref{tab:semantics} makes this attribution causal in direction, since removing the self-distillation objective removes the semantic gain while keeping it improves fidelity at the same time, and the cross-interface experiments extend it to objective transfer, since the same recipe trains successfully on both auxiliary interfaces and the latent-to-latent model reproduces the prefix-reliability trend.

\subsection{Limitations}

We further discuss limitations for concrete openings. Because capacity is counted in continuous scalars rather than entropy-coded bits, quantizing the ordered sequence would turn prefix decoding into a genuine variable-rate codec and place semantic structure on a rate--distortion frontier. Because our semantic evidence is global and classification-based, causal token interventions and dense prediction would establish what individual positions encode. Most importantly, better reconstruction and frozen semantics did not produce better generation, which isolates latent distribution geometry rather than autoencoder quality as the binding constraint on compact tokenizers, and raises the question of how much semantic structure an EMA teacher can organize before an external target becomes necessary.

\section{Conclusion}

In this paper we present \method{}, a compact visual autoencoder that learns its semantic targets from the same generative system that must reconstruct the image. We find that a pretrained recognition encoder is unnecessary for semantic organization: an EMA copy of the autoencoder itself supplies targets that raise frozen linear probing by 5.1 points over a reconstruction-only baseline while reconstruction improves at the same time. We also find that position-dependent token corruption orders information along the sequence, so a single trained model decodes variable-length prefixes without retraining. Overall, \method{} improves all four reconstruction metrics over the closest representation-pivoted baseline at half its scalar capacity, and its 128-token prefix still exceeds that baseline in PSNR and SSIM at one quarter. We hope our work provides useful insights for building semantic tokenizers without external supervision.

\bibliography{withinvae}
\bibliographystyle{iclr2027_conference}

\end{document}